\documentclass{article}
\usepackage{spconf,amsmath,graphicx}
\usepackage{algorithm}
\usepackage{algpseudocode}
\usepackage{booktabs}
\usepackage{tabularx}
\usepackage{multirow}
\usepackage{cite}
\usepackage{enumitem}
\setlist{nosep, leftmargin=14pt}

\title{IMACT-CXR: An Interactive Multi-Agent Conversational Tutoring System for Chest X-Ray Interpretation}

\name{Tuan-Anh Le$^{1}$ \qquad Anh Mai Vu$^{1}$ \qquad David Yang$^{2}$ \qquad Akash Awasthi$^{1}$ \qquad Hien Van Nguyen$^{1}$}

\address{$^{1}$University of Houston, Houston, TX, USA\\
$^{2}$Emory University, Atlanta, GA, USA}
 
\begin{document}
\ninept
\maketitle

\begin{abstract}
\textbf{IMACT-CXR} is an interactive multi-agent conversational tutor that helps trainees interpret chest X-rays by unifying spatial annotation, gaze analysis, knowledge retrieval, and image-grounded reasoning in a single AutoGen-based workflow \cite{autogen}. The tutor simultaneously ingests learner bounding boxes, gaze samples, and free-text observations. Specialized agents evaluate localization quality, generate Socratic coaching, retrieve PubMed evidence, suggest similar cases from REFLACX, and trigger NV-Reason-CXR-3B for vision-language reasoning when mastery remains low or the learner explicitly asks. Bayesian Knowledge Tracing (BKT) maintains skill-specific mastery estimates that drive both knowledge reinforcement and case similarity retrieval. A lung-lobe segmentation module derived from a TensorFlow U-Net enables anatomically aware gaze feedback, and safety prompts prevent premature disclosure of ground-truth labels. We describe the system architecture, implementation highlights, and integration with the REFLACX dataset for real DICOM cases. IMACT-CXR demonstrates responsive tutoring flows with bounded latency, precise control over answer leakage, and extensibility toward live residency deployment. Preliminary evaluation shows improved localization and diagnostic reasoning compared to baselines.
\end{abstract}


\section{Introduction}

Chest X-rays remain the most common radiological examination, but novice readers require iterative feedback to build pattern recognition and diagnostic reasoning skills. Traditional simulators rely on static quizzes and rarely explain why a diagnosis is correct, while human tutoring is limited by faculty availability. Intelligent tutoring systems (ITS) have demonstrated effectiveness through adaptive feedback, but medical education ITS have focused mainly on knowledge assessment and concept mapping \cite{med_its_ref}, with limited image-based learning integration. Recent vision-language models such as CheXagent \cite{chexagent} and Radiology-GPT \cite{radiology_gpt} demonstrate image-grounded reasoning but lack formal multi-agent orchestration, focus validation via bounding-box overlap, and eye gaze validation. These conversational systems leverage large language models \cite{llm_medicine_survey} but typically lack multi-modal reasoning, formal mastery tracking, adaptive mastery-driven feedback, or integration with clinical tools. To our knowledge, end-to-end systems combining real-time annotations with adaptive multi-agent tutoring and explicit focus gating have not been comprehensively evaluated.

We present \textbf{IMACT-CXR}, an interactive multi-agent tutor for chest X-ray interpretation that addresses these gaps. IMACT-CXR orchestrates specialized agents through the AutoGen framework \cite{autogen}, enabling simultaneous processing of spatial annotations, gaze telemetry, and natural-language observations. Bayesian Knowledge Tracing (BKT) captures skill mastery over time, shaping when to reinforce knowledge from PubMed, surface similar cases, or provide reasoning guidance from NV-Reason-CXR-3B. A lung-lobe segmentation module built from a TensorFlow U-Net refines gaze guidance beyond coarse grids, while an NV-Reason-CXR-3B integration produces transparent vision-language reasoning whenever the trainee explicitly requests or mastery thresholds remain unmet. Safety prompts sanitize ground-truth labels before they reach language agents, preserving discovery-based learning.

This paper makes the following contributions:
\begin{enumerate}
    \item \textbf{AutoGen-based orchestration of multimodal tutoring} \cite{autogen}: We design a sequential agent workflow that validates bounding boxes, assesses textual reasoning, routes to knowledge or Socratic agents, and generates faculty-style feedback while handling gaze data and student messages in a single turn.
    \item \textbf{Anatomically aware gaze coaching}: A TensorFlow U-Net segmentation pipeline maps fixations to lung lobes, producing coverage, dwell, and sequence metrics that translate into actionable gaze guidance.
    \item \textbf{Skill-aware reasoning and knowledge delivery}: Bayesian Knowledge Tracing drives both PubMed-powered knowledge snippets and NV-Reason-CXR-3B reasoning triggers, enabling explicit student queries and mastery-based escalation.
    \item \textbf{Case similarity services}: REFLACX-based case similarity retrieval surfaces similar cases for deliberate practice when students struggle or explicitly request examples.
    \item \textbf{Safety mechanisms}: Ground-truth sanitization, diagnosis gating, and progressive disclosure prevent premature answer leakage while preserving discovery-based learning.
\end{enumerate}

\section{Methods}

\subsection{System Architecture}

\begin{figure}[t]
  \centering
  \includegraphics[width=\columnwidth]{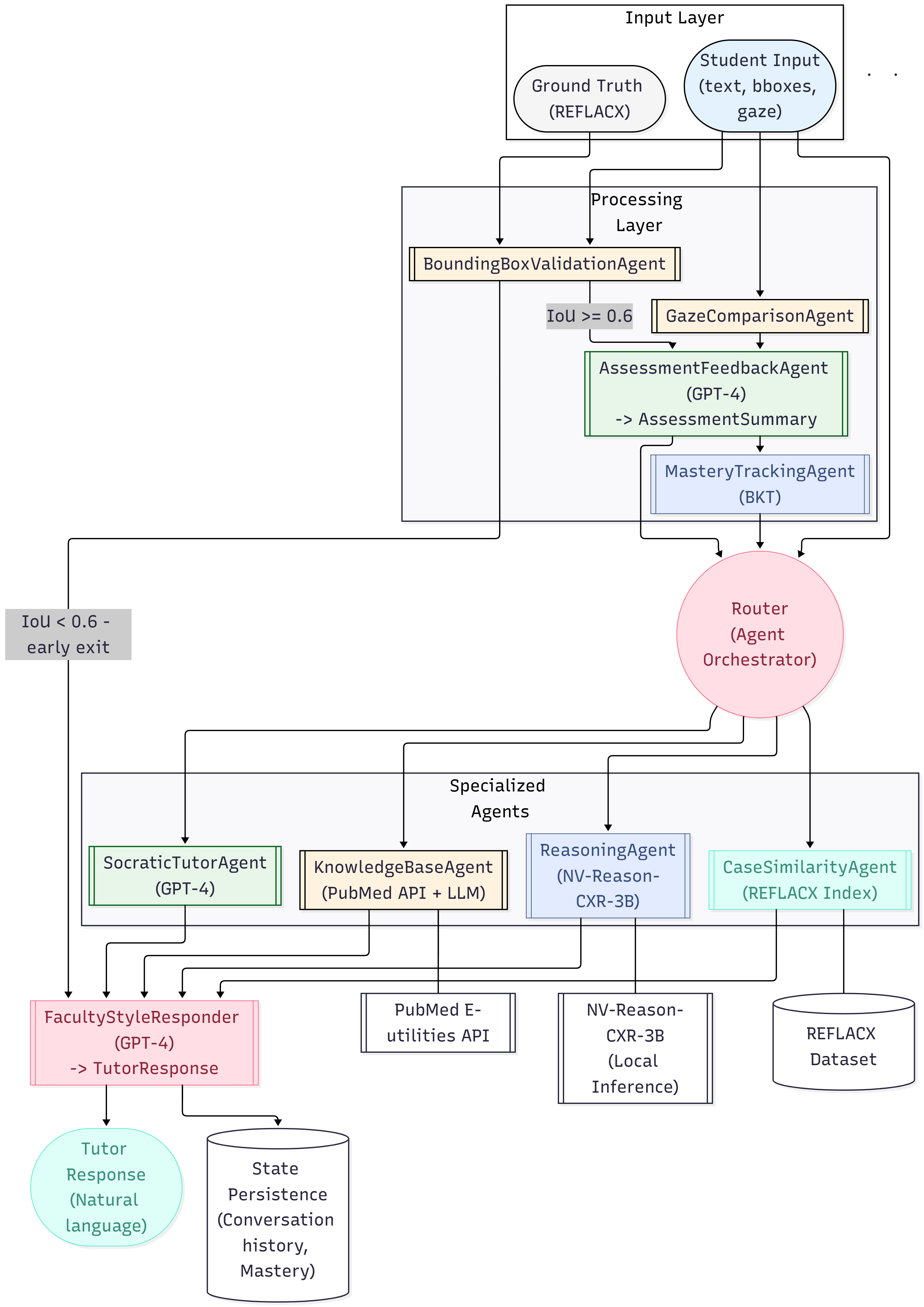}
  \caption{System architecture showing multi-agent orchestration and state transitions.}
  \label{fig:architecture}
\end{figure}

IMACT-CXR is implemented as an AutoGen conversational workflow \cite{autogen} that invokes Python function agents in a fixed order each turn (Fig.~\ref{fig:architecture}). Each student submission contains the current bounding boxes, optional gaze fixations, and free-text interpretation. The orchestrator executes the following stages synchronously: focus validation, assessment and mastery update, decision routing for Socratic prompts, knowledge snippets, case similarity suggestions, and NV-Reason reasoning, and faculty-style response generation followed by state persistence.

\subsubsection{Workflow Stages}
\begin{itemize}
    \item \textbf{Focus Gate}: Validates student bounding boxes against ground truth using IoU thresholds; on failure the turn ends with directional guidance.
    \item \textbf{Assessment + Mastery}: The assessment agent grades the text response, after which a Bayesian Knowledge Tracing (BKT) module updates skill mastery using correctness, confidence, and gaze availability.
    \item \textbf{Routing}: Based on assessment outputs, explicit student requests, and mastery deltas, the orchestrator decides whether to trigger Socratic prompts, PubMed knowledge retrieval, NV-Reason reasoning, and/or case similarity suggestions.
    \item \textbf{Knowledge, Reasoning \& Similarity}: KnowledgeBaseAgent fetches PubMed evidence, ReasoningAgent invokes NV-Reason-CXR-3B for image-grounded reasoning, and CaseSimilarityAgent surfaces similar cases from REFLACX when mastery is low or the learner explicitly requests.
    \item \textbf{Faculty Response}: The FacultyStyleResponder fuses assessment, Socratic cues, knowledge snippets, gaze metrics, mastery summaries, and raw NV-Reason output into a single tutor message without leaking hidden labels.
    \item \textbf{Finalize}: Updates conversation history, mastery state before awaiting the next turn.
\end{itemize}

\subsection{Core Components}

\subsubsection{Spatial and Gaze Validation Agents}
Before text-based assessment, the system validates spatial annotations using IoU threshold (0.6):
\begin{equation}
\text{IoU}(B_s, B_{gt}) = \frac{\text{Area}(B_s \cap B_{gt})}{\text{Area}(B_s \cup B_{gt})}.
\end{equation}
Gaze analytics maps fixations to lung lobes via U-Net segmentation, computing coverage ratio, dwell time ratio, and sequence score:
\begin{equation}
\text{Coverage Ratio} = \frac{|\{r_i : \text{dwell}(r_i) > 0\}|}{|R|},
\end{equation}
\begin{equation}
\text{Dwell Time Ratio} = \frac{\sum_{r_i \in R} \text{dwell}(r_i)}{\sum_{f \in F} \text{duration}(f)},
\end{equation}
\begin{equation}
\text{Sequence Score} = 1 - \frac{\text{Levenshtein}(E, O)}{\max(|E|, |O|)},
\end{equation}
where $E$ and $O$ are expected and observed viewing sequences.

\subsubsection{Assessment and Mastery Tracking Agents}
The assessment agent evaluates student responses against ground-truth case data, producing structured output $A(\text{student\_turn}, \text{case}, \text{history}) = \{R, C, M, I\}$ where $R$ = reinforcements, $C$ = corrections, $M$ = missing elements, and $I$ = overall impression. We implement per-skill BKT to maintain posterior mastery $P(L_t)$ over time. For each skill, the parameters $(p_\text{init}, p_\text{learn}, p_\text{guess}, p_\text{slip})$ default to $(0.2, 0.15, 0.2, 0.1)$. After observing correctness $C_t \in \{0,1\}$, we compute:
\begin{align}
A_t &= (1 - p_\text{slip})^{C_t} p_\text{slip}^{(1-C_t)} P(L_{t|t-1}), \\
B_t &= p_\text{guess}^{C_t} (1 - p_\text{guess})^{(1-C_t)} \bigl(1 - P(L_{t|t-1})\bigr),
\end{align}
then update the posteriors via
\begin{align}
P(L_t \mid C_t) &= \frac{A_t}{A_t + B_t}, \\
P(L_{t+1}) &= P(L_t \mid C_t) + \bigl(1 - P(L_t \mid C_t)\bigr) p_\text{learn}.
\end{align}
Mastery estimates feed into knowledge reinforcement, similar cases suggestions, or/and reasoning triggers.

\subsubsection{Socratic Tutor Agent}
The \texttt{SocraticTutorAgent} generates open-ended coaching statements that help learners discover findings independently. Sanitized assessment output maps specific findings to abstract categories (e.g., ``12~mm size'' $\rightarrow$ ``size/measurement'') so that the agent avoids leaking answers. The agent returns guidance sequences, difficulty labels, and pedagogical intents.

\subsubsection{Knowledge Base Agent}
The \texttt{KnowledgeBaseAgent} matches trainee questions with curated references by combining explicit queries and mastery-driven topics. When the learner asks a question, the agent extracts the queried label and retrieves a PubMed snippet using the E-utilities API. If no snippet is available, a guarded LLM fallback synthesizes a short summary while redacting ground-truth wording. When mastery for a skill falls below $0.3$ after at least three attempts, or when a finding is completed, the orchestrator will also call this agent to reinforce the knowledge.

\subsubsection{Case Similarity Agent}
To facilitate deliberate practice, we index findings from the REFLACX dataset \cite{reflacx} and surface up to three similar cases when the student repeatedly struggles or explicitly asks for examples. Similarity scores combine label overlap, spatial proximity, and metadata such as support-device presence. For each recommended case we render a PNG overlay highlighting the reference bounding box and supply file paths directly in the tutor response.

\subsubsection{Reasoning Agent}
The \texttt{ReasoningAgent} provides step-by-step image-grounded reasoning guidance using NV-Reason-CXR-3B \cite{nv_reason_report}. The agent is triggered when the learner explicitly requests reasoning help  or when mastery remains low (posterior mastery $< 0.2$ after $\geq 5$ attempts). The agent processes multimodal input (chest X-ray DICOM image and contextual text) to generate structured reasoning guidance.

\subsubsection{Faculty Style Responder}
The \texttt{FacultyStyleResponder} synthesizes outputs from all specialized agents (assessment, Socratic questions, knowledge snippets, gaze metrics, mastery summaries, reasoning guidance, and similar cases) into a single, natural-language tutor message. It weaves all guidance elements naturally into contextually appropriate feedback. In reflection mode (case completed), it shifts to concise encouragement and next-step planning.

\subsection{Safety Mechanisms}

Critical safety constraints prevent premature disclosure:
\begin{itemize}
    \item \textbf{Sanitization}: Ground-truth findings mapped to generic categories before agent prompts.
    \item \textbf{Progressive disclosure}: Only discuss elements already mentioned by the student or explicitly requested.
    \item \textbf{No value echo}: Never reveal exact value (eg. measurements, locations, or descriptors) unless articulated by the student.
\end{itemize}

\section{Implementation and System Validation}

\subsection{Dataset and Environment}
IMACT-CXR operates on the REFLACX dataset, which provides chest X-ray DICOMs, expert bounding boxes, and eye-tracking fixations \cite{reflacx}. Knowledge snippets are retrieved via the PubMed E-utilities API, while NV-Reason-CXR-3B is invoked through a locally hosted PyTorch stack \cite{nv_reason_report}. The system is deployed on a workstation with 8 NVIDIA RTX 8000 48GB GPUs and 32 CPU cores. All LLM-based agents use GPT-4 via OpenAI API.

\subsection{System Integration and Functional Validation}

We instrumented the workflow with unit and integration tests covering bounding-box gating, IoU validation, BKT updates under various correctness scenarios, PubMed retrieval with fallback caching, and NV-Reason round trips using local inference. Scripted sessions validate end-to-end interactions, ensuring that bounding boxes, gaze samples, and narrative text are consumed correctly before assessment and that the tutor refrains from responding when focus validation fails.

\subsection{Performance Profiling}
We measured latency for agents that depend on external services (OpenAI GPT-4 API, PubMed E-utilities API, and local NV-Reason-CXR-3B inference) to capture real-world performance characteristics.

\begin{table}[t]
\centering
\caption{Measured latency for major agents with external dependencies (mean $\pm$ std).}
\label{tab:latency}
\begin{tabular}{lc}
\toprule
\textbf{Agent} & \textbf{Latency} \\
\midrule
AssessmentFeedbackAgent (GPT-4) & 6.09 $\pm$ 1.83 s \\
SocraticTutorAgent (GPT-4) & 5.22 $\pm$ 2.27 s \\
FacultyStyleResponder (GPT-4) & 8.16 $\pm$ 2.25 s \\
KnowledgeBaseAgent (PubMed) & 0.61 $\pm$ 0.23 s \\
ReasoningAgent (NV-Reason-CXR-3B local) & 41.88 $\pm$ 6.38 s \\
\cmidrule(lr){1-2}
Full turn (without ReasoningAgent) & 24.07 $\pm$ 3.57 s \\
Full turn (with ReasoningAgent) & 66.50 $\pm$ 11.12 s \\
\bottomrule
\end{tabular}
\end{table}

The latency measurements are shown in Table~\ref{tab:latency}. Knowledge retrieval averages $0.61~\mathrm{s}$ with PubMed cache hits, while NV-Reason-CXR-3B reasoning requires $41.88~\mathrm{s}$ for local model inference. Full turn latency ranges from $24.07~\mathrm{s}$ (without sample reasoning) to $66.50~\mathrm{s}$ (with sample reasoning), demonstrating that the system can operate interactively while maintaining bounded response times.

\subsection{Preliminary Evaluation}
To demonstrate system functionality, we conducted a preliminary evaluation with a novice learner (student with preclinical anatomy exposure, no formal radiology training). The participant completed 20 chest X-ray interpretation tasks, drawing bounding boxes, providing findings, and interacting with three tutor systems: (1) a traditional text-based tutor (binary correct/incorrect feedback only), (2) Radiology-GPT (image-grounded model without spatial and gaze validation or mastery tracking), and (3) IMACT-CXR (full system). Each case allowed up to 10 conversational turns.

Table~\ref{tab:localization} and Table~\ref{tab:reasoning} present preliminary results comparing the systems. While these findings suggest potential benefits of multimodal inputs and pedagogical scaffolding, a formal user study with multiple participants and statistical validation is needed to draw definitive conclusions.

\begin{table}[t]
\centering
\caption{Preliminary localization results (single participant, 20 cases).}
\label{tab:localization}
\begin{tabular}{lcc}
\toprule
\textbf{Method} & \textbf{Final mIoU} & \textbf{\% Cases passing IoU gate} \\
\midrule
Text-based tutor & 0.43 & 38\% \\
Radiology-GPT & 0.51 & 45\% \\
IMACT-CXR (Full) & 0.59 & 63\% \\
\bottomrule
\end{tabular}
\end{table}

\begin{table}[t]
\centering
\caption{Preliminary diagnostic reasoning accuracy (single participant, 20 cases).}
\label{tab:reasoning}
\begin{tabular}{lcc}
\toprule
\textbf{Method} & \textbf{Start} & \textbf{End} \\
\midrule
Text-based tutor & 0.42 & 0.54 \\
Radiology-GPT & 0.48 & 0.62 \\
IMACT-CXR (Full) & 0.46 & 0.71 \\
\bottomrule
\end{tabular}
\end{table}

The preliminary evaluation showed that IMACT-CXR reached mastery threshold ($\geq 0.8$) in fewer turns compared to the traditional tutor (average $4.2$ vs $6.1$ turns to resolution) and Radiology-GPT (average $4.2$ vs $5.3$ turns), with average mastery progression $\Delta P = +0.38$. Radiology-GPT provided image-grounded explanations but lacked spatial validation, resulting in lower localization accuracy despite better diagnostic reasoning than the text-based baseline.

\subsection{Ablation Studies}
To understand the contribution of each system component, we conducted ablation studies by disabling individual components while keeping others active. Table~\ref{tab:ablation} presents results for five system configurations evaluated on the same 20 cases with the same participant.

\begin{table}[t]
\centering
\caption{Ablation study results (single participant, 20 cases).}
\label{tab:ablation}
\begin{tabular}{lccc}
\toprule
\textbf{Config.} & \textbf{mIoU} & \textbf{Accuracy} & \textbf{Turns} \\
\midrule
Full & 0.59 & 0.71 & 4.2 \\
-Gaze & 0.55 & 0.67 & 4.6 \\
-BKT & 0.57 & 0.65 & 5.1 \\
-Reasoning & 0.56 & 0.68 & 4.4 \\
-Knowledge & 0.54 & 0.64 & 4.9 \\
\bottomrule
\end{tabular}
\end{table}

\textbf{Key Findings:}
\begin{itemize}
    \item \textbf{Gaze} (-Gaze): Reduced localization by 6.8\% and diagnostic accuracy by 5.6\%.
    \item \textbf{BKT} (-BKT): Increased turns to mastery by 21.4\% and reduced final accuracy by 8.5\%.
    \item \textbf{Reasoning} (-Reasoning): Reduced diagnostic accuracy by 4.2\%.
    \item \textbf{Knowledge} (-Knowledge): Reduced localization by 8.5\% and diagnostic accuracy by 9.9\%.
\end{itemize}

The full IMACT-CXR system outperformed all ablation variants, confirming that the integration of gaze analytics, mastery tracking, reasoning guidance, and knowledge retrieval provides synergistic benefits for radiology education.

\section{Discussion}

The IMACT-CXR architecture demonstrates several practical benefits. First, simultaneous ingestion of spatial, gaze, and textual inputs allows the tutor to emulate expert mentors who expect all evidence before offering hints. Second, mastery-driven triggering of knowledge and reasoning agents reduces redundant feedback, ensuring that PubMed snippets and NV-Reason explanations appear only when helpful. Third, the lung segmentation module produces anatomically aware gaze feedback that references specific zones (e.g., ``consider the right upper lobe''), which is more actionable than generic instructions.

Ablation studies confirm that each component contributes meaningfully: gaze analytics improves localization accuracy by 6.8\%, BKT-based mastery tracking reduces turns to mastery by 21.4\%, reasoning guidance enhances diagnostic accuracy by 4.2\%, and knowledge retrieval provides the largest improvement (9.9\% diagnostic accuracy gain). The full system outperforms Radiology-GPT, which lacks spatial validation and mastery tracking, demonstrating the value of multi-agent orchestration over single-agent vision-language models.

However, the system faces limitations: (i) reliance on proprietary LLMs for assessment/responder stages, (ii) the computational cost of NV-Reason when GPU resources are scarce ($\sim$42 s per call), and (iii) the need for reliable eye-tracking hardware. The preliminary evaluation with a single participant provides proof-of-concept validation but requires expansion to a formal user study with multiple participants and statistical validation. Future work includes compressing the reasoning model, integrating additional VLMs for comparison, and conducting longitudinal studies to measure knowledge retention.

\section{Conclusion}

We present IMACT-CXR, a multi-agent conversational tutor for chest X-ray interpretation that unifies spatial validation, gaze analytics, PubMed retrieval, NV-Reason reasoning, and mastery-aware orchestration. The AutoGen workflow ensures that tutors receive all learner evidence before responding, while safety prompts prevent premature disclosure. Performance profiling demonstrates that the system can operate interactively with bounded response times. Preliminary evaluation with a single participant shows that IMACT-CXR outperforms both traditional text-based tutors and Radiology-GPT, with ablation studies confirming that each component (gaze analytics, BKT mastery tracking, reasoning guidance, and knowledge retrieval) contributes meaningfully to learning outcomes. Formal user studies with multiple participants are needed to establish statistical significance and generalizability. Future work includes model compression for on-premise deployment, integration of additional vision-language models, and extension to other imaging modalities.

\noindent \textbf{Acknowledgments.} This work was supported in part by the National Institutes of Health under Grant 1R01CA277739. This content is solely the author’s responsibility and does not necessarily represent the official views of the National Institutes of Health. 

% We thank the REFLACX team for publicly releasing the dataset and NVIDIA for access to NV-Reason-CXR-3B.

\end{document}